\documentclass[letterpaper, 10 pt, conference]{ieeeconf}
\IEEEoverridecommandlockouts                              

\usepackage{graphicx} 
\usepackage{amsmath} 
\usepackage{amssymb}  
\usepackage{float}
\usepackage{ifthen}
\usepackage[ruled,linesnumbered]{algorithm2e}

\SetCommentSty{mycommfont}
\usepackage{tabularx}
\newcolumntype{b}{>{\hsize=.9\hsize}X}

\usepackage[sort,compress]{cite}

\usepackage{url}

\usepackage{tikz}
\usepackage{textcomp}
\usepackage{hyperref}
\usepackage{subcaption}

\newboolean{arxiv}
\setboolean{arxiv}{true}

\ifthenelse{\boolean{arxiv}}{
\def\doi{DOI: \href{https://doi.org/XX.YYYYY/X.YYYYY}{XX.YYYYY/X.YYYYY}}

\newcommand\copyrightnotice{%
\begin{tikzpicture}[remember picture,overlay]%
\node[anchor=south,yshift=10pt] at (current page.south) {\fbox{\parbox{\dimexpr\textwidth-\fboxsep-\fboxrule\relax}{\copyrighttext}}};%
\end{tikzpicture}%
}
}{
\newcommand\copyrightnotice{}
}

\newcommand{\argmax}{\operatornamewithlimits{argmax}}
\newcommand{\argmin}{\operatornamewithlimits{argmin}}

\newenvironment{myAlgorithm}[1]{\begin{algorithm}[ht]\DontPrintSemicolon\caption{#1}}{\end{algorithm}}
\SetKwProg{myCodeBlock}{}{}{}
\newcommand{\algoIndent}[1]{\myCodeBlock{}{#1}}

\makeatother

\title{\LARGE \bf
Active Reward Learning for Co-Robotic Vision Based Exploration in Bandwidth Limited Environments
}

\author{Stewart Jamieson$^{1,2}$, Jonathan P. How$^{2}$, Yogesh Girdhar$^{3}$
\thanks{*This work was supported by NSF-NRI Award Number 1734400}
\thanks{$^{1}$S. Jamieson is with the MIT-WHOI Joint Program in Applied Ocean Science and Engineering
        {\tt\small sjamieson@whoi.edu }}%
\thanks{$^{2}$S. Jamieson and J.P. How are with the Department of Aeronautics and Astronautics at the Massachusetts Institute of Technology (MIT)
        {\tt\small \{sjamieson,jhow\}@mit.edu }}%
\thanks{$^{3}$Y. Girdhar is with the Applied Ocean Physics and Engineering Department at the Woods Hole Oceanographic Institution (WHOI)
        {\tt\small yogi@whoi.edu}}%
}

\begin{document}
\maketitle
\thispagestyle{empty}
\pagestyle{empty}

\begin{abstract}
We present a novel POMDP problem formulation for a robot that must autonomously decide where to go to collect new and scientifically relevant images given a limited ability to communicate with its human operator. From this formulation we derive constraints and design principles for the observation model, reward model, and communication strategy of such a robot, exploring techniques to deal with the very high-dimensional observation space and scarcity of relevant training data. We introduce a novel active reward learning strategy based on making queries to help the robot minimize path ``regret'' online, and evaluate it for suitability in autonomous visual exploration through simulations. We demonstrate that, in some bandwidth-limited environments, this novel regret-based criterion enables the robotic explorer to collect up to 17\% more reward per mission than the next-best criterion.
\end{abstract}


\section{Introduction}

\copyrightnotice%
Images of exotic biological and geological phenomena from remote and dangerous locations have tremendous scientific value but are extraordinarily challenging and costly to collect. Robots have been at the forefront of collecting visual scientific observations in such environments, which include Mars \cite{Estlin2012}, deep space \cite{Gao2017}, the Earth's oceans \cite{Ballard1993,Foley2009,Clarke2009}, and under Arctic ice sheets \cite{Williams2015}. Communication bandwidth constraints are perhaps the biggest bottleneck to exploration in these remote environments~\cite{Kaeli2014,Burroughes2016}. As such, common current approaches to autonomous exploration are to either deploy the vehicles on a predefined path or to deploy them with adaptive path plans based on tracking low-dimensional observations from some other sensor. This paper proposes a novel approach to vision-guided exploration using a human-robot team that is effective even in the presence of strong bandwidth constraints such as those imposed by acoustic underwater communications~\cite{Kaeli2014}.

\begin{figure}[ht]
     \centering
     \includegraphics[width=.96\linewidth]{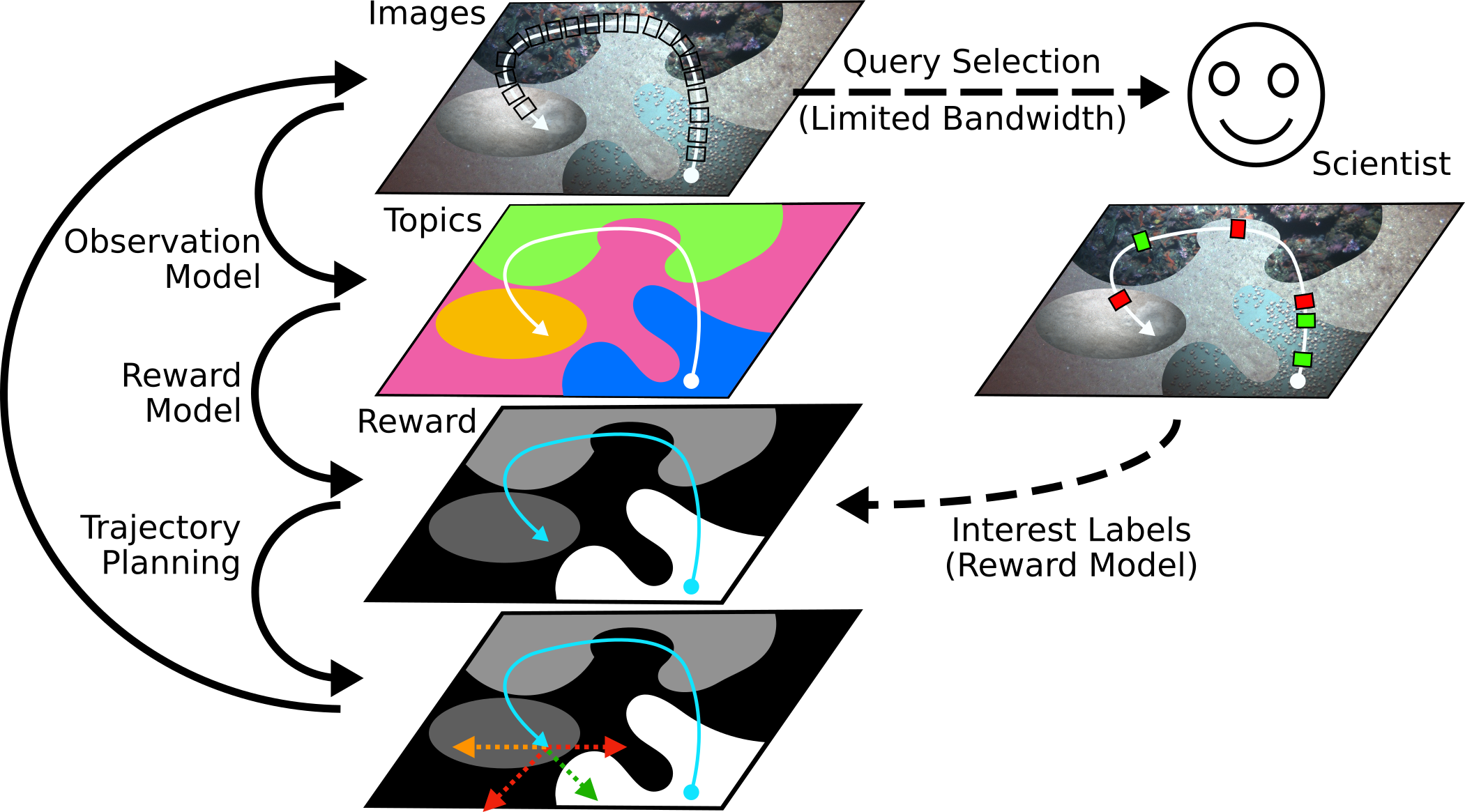}
     \caption{Proposed approach to co-robotic exploration that models the interest of the operator over a low bandwidth communication channel and uses the learned reward model to plan the most rewarding (in terms of interest) robot paths. }
     \label{fig:overview}
\end{figure}


Most recent progress towards increasing the science return of autonomous exploration missions has been made through enabling ``opportunistic science''\footnote{This refers to an autonomous action, such as targeting and using a particular sensor, that preempts the robot's current task after being triggered by a specific phenomena recognized by an onboard detection algorithm.} as well as addressing challenges in navigation, task planning and scheduling autonomy \cite{Chien2005,Estlin2007,Castano2006}. The progress in 
adaptive sampling and exploration algorithms for robots has primarily focused on observing spatially-varying \textit{scalar} quantities, such as temperature \cite{Hitz2017,Flaspohler2019}. To reach a future of efficient robotic explorers in remote environments, robots will need to autonomously: recognize visual phenomena that might be scientifically interesting, transmit images of them to scientists for clarification as needed, model where more of them might be found, and plan a trajectory accordingly (Figure \ref{fig:overview}).

The primary contributions of this work are a partially observable Markov decision process (POMDP) formulation for vision-based scientific exploration and a solution that is generalizable to many environments. Note that we explicitly constrain the focus of this work to dealing with high-dimensional observation spaces when solving the POMDP. 
The proposed exploration approach uses the limited communication bandwidth to query the operator for the \textit{value} of representative images and uses the responses to learn an \textit{interest function} that informs the robot about the value of an exploration path. 
The proposed approach is suitable for deployment in completely unknown environments, and it can use (but does not require) prior knowledge about the environment and the phenomena being observed. 
Our final contribution is an analysis and comparison of active learning decision criteria that a robot could use for deciding which observations to send to the operator. That analysis is supported by simulations of a scientific exploration task using both real and artificial data.



\section{Related Work}

This work contributes to the field of \textit{autonomous science}; previous works in this area include AEGIS \cite{Estlin2012} and OASIS \cite{Castano2006}, which enabled robots to opportunistically recognize scientifically relevant image observations, given a predefined model, and schedule more detailed observations with other sensors. However, these algorithms required domain-specific feature engineering and lacked spatial observation models, so the adaptive path planning was limited to moving the robot closer to a target that had already been detected. At the other extreme, ``curious'' robots use a generic unsupervised vision model and autonomously move towards anything in their environment that is surprising or novel to the model \cite{Girdhar2016a}; the lack of operator input makes it impossible to directly specify particular scientific objectives using this approach.

Our work is closely related to the work of Arora et al.~\cite{Arora2017}, which modeled operator's domain knowledge with a pre-defined Bayesian Network (BN) that was used by the robot to estimate the reward for a trajectory. They introduced a spatial observation model in the system, enabling informative path planning using Monte-Carlo Tree Search (MCTS) to explore an action tree composed of movement and sensing actions \cite{Arora2017}. Their approach requires the operator to specify the domain-specific BN \textit{a priori}, and has limited utility as a general purpose exploration tool that can be deployed in unknown environments. In contrast, our solution learns the reward model online, and hence allows the robot to deal with unexpected observations efficiently during exploration.

\textit{Active learning} algorithms interactively query an oracle to produce samples in the training set, such that the model can be trained with far fewer labeled examples than would normally be required~\cite{Balcan2010}. Active reward learning algorithms have efficiently learned reward models representing human ratings or preferences for robot behaviours by making on the order of 10-100 reward queries~\cite{Daniel2014,Sadigh2017}. Doshi-Velez et al.~\cite{Doshi-Velez2012} considered a query to be an action that could be taken if it helped the robot to gain additional reward. This is \textit{online} active learning and our approach is most related to theirs with the main difference that we use reward queries to learn a mapping from observations to reward, whereas~\cite{Doshi-Velez2012} used policy queries to learn optimal actions directly.

Due to the high-dimensionality of natural images, even with active learning, it can take hundreds of queries to learn a reward model~\cite{Shkurti2018}. In bandwidth limited environments such as the deep sea, sending that many images for labelling during the span of a mission is not feasible. Deep features are relatively low-dimensional representations of images which are very helpful for learning new classification tasks with few examples~\cite{Donahue14,Romero2015}. Topic models, especially when combined with deep features, can be used to provide a low-dimensional semantic representation of the visual environment~\cite{Flaspohler2017}. Our proposed POMDP approach leverages both active learning and low-dimensional image representations to enable interactive visual exploration over low bandwidth.


\section{The Co-Robotic Visual Exploration POMDP}

We present the co-robotic visual exploration problem as a POMDP. We model the state of the robot
at time $t$ as $S_{t}=\left(X_{t},Y_{t},L_{t}\right)$. $X_{t}=\left\{ \left(\boldsymbol{x}_{i},I_{i}\right)\right\} _{i=1}^{t}$
is the sequence of locations the robot has visited, with corresponding
image observations, where the current location is $\boldsymbol{x}_{t}\in\mathbb{R}^{3}$ and the latest observation is the image $I_t \in \mathbb{O}$. $L_{t}$ is the set of indices of images sent to and labeled by the operator. $Y_t$ contains the reward labels for all images, including those that have not been sent; most of these are unknown, making the robot's state partially observable.

The partial observability comes from the robot's
limited ability to query the operator during a mission; in bandwidth constrained environments the robot sends images at a much slower rate than it collects them, so it must decide which labels to observe. We assume that only the operator can evaluate
the unknown, but deterministic, binary ``interest'' function $\mathcal{I}\left(I\right)$
such that $\left(Y_{t}\right)_{i}=\mathcal{I}\left(I_{i}\right)$.
Further, it is assumed that the operator cannot express their interest function analytically (otherwise it would be computed onboard the robot), and would instead train an approximate model based on their labels for various example images. 
However, since exploration typically occurs in remote and unstudied environments, the operator does not have a fully representative dataset of what the robot will observe and is unable to provide the robot with a complete model of $\mathcal{I}\left(\cdot\right)$
\emph{a priori}. 

The entire POMDP is characterized by the tuple $\left(\mathcal{S},\mathcal{A},\mathbb{O},T,O,R,\gamma,b_{0}\right)$:
\begin{table}[H] \vspace*{-.1in}
\centering
\begin{tabularx}{\linewidth}{|c|X|X|}
\hline 
Component & Definition & Our Assumptions\tabularnewline
\hline 
\hline 
$\mathcal{S}$ & State space of the robot & $S=\left(X,Y, L\right)$ \tabularnewline
\hline 
$\mathcal{A}$ & Discrete set of robot actions & Motion primitives\footnotemark \tabularnewline
\hline 
$\mathbb{O} $ & Observation space & Natural images \newline and binary labels \tabularnewline
\hline 
$T$ & Transition function & $\mathcal{S}\times\mathcal{A}\mapsto\mathcal{S}$ \tabularnewline
\hline 
$O$ & Observation model & $\mathcal{S}\times\mathcal{A}\mapsto\mathbb{O}$ \tabularnewline
\hline 
$R$ & Reward model & $\mathcal{S}\times\mathcal{A}\times \mathbb{O}\mapsto \mathbb{R}$ \tabularnewline
\hline 
$\gamma $ & Discount factor & $\gamma \in [0,1]$ \tabularnewline
\hline 
$b_0$ & Initial belief state & Initial location $\boldsymbol{x}_0$ \tabularnewline
\hline 
\end{tabularx}
\end{table}
\footnotetext{Querying the operator is often modelled (e.g., in~\cite{Doshi-Velez2012}) as another action in $\mathcal{A}$ with some cost of communication, such as energy usage, included in the reward model. For simplicity, we assume this cost is negligible and that the robot performs queries concurrently with other actions.}
Given these specifications, it is typical for the robot to use an online POMDP planner to approximate an optimal
policy $\pi^{\star}:\mathcal{S}\mapsto\mathcal{A}$ in real-time. Algorithm \ref{alg:corobotic-exploration}  presents our approach to co-robotic exploration based on the assumptions listed above.

There are three key decisions to fully specify the co-robotic visual
exploration POMDP that we will consider. The first is defining an observation model
over the space of natural images. The
second is defining a reward model, and the
third is choosing an effective active learning strategy. 

\begin{myAlgorithm}{Co-Robotic Exploration}
\label{alg:corobotic-exploration}
\textbf{Given:} $\left(\mathcal{S},\mathcal{A},\mathbb{O},T,\mathcal{O},R,\gamma,\boldsymbol{x}_{0}\right)$, $t_{\text{max}}$

$X_{0}\leftarrow\varnothing$  \tcp*{Stores the path and observations}

$\boldsymbol{\tau} \leftarrow \{\boldsymbol{x}_0\}$  \tcp*{The current trajectory plan}

$q\leftarrow$ \emph{null} \tcp*{Index of next observation to label}

$t\leftarrow1$ \tcp*{The current timestep}

\textbf{while} $t<t_{\text{max}}$:\algoIndent{
$\boldsymbol{x}_{t}\leftarrow\textsc{Next\_Step}(\boldsymbol{\tau})$

$I_{t}\leftarrow$ \textsc{Observe}(\textbf{$\boldsymbol{x}_{t}$})

$X_{t}\leftarrow X_{t-1}\cup\left\{ \boldsymbol{x}_{t},I_{t}\right\} $

$\mathcal{O}\leftarrow$\textsc{ Update\_Observation\_Model}($\mathcal{O},X_{t}$)

\textbf{if} \textsc{Label\_Ready}($I_{q}$)\algoIndent{$y_{q}\leftarrow$ \textsc{Query\_Result}($I_{q}$)

$Y_{t}\leftarrow Y_{t-1}\cup\left\{ I_{q},y_{q}\right\} $

$R\leftarrow$\textsc{ Update\_Reward\_Model}($R,Y_{t}$)

$q\leftarrow$ \emph{null}}

\textbf{endif}

$\boldsymbol{\tau}\leftarrow$ \textsc{Plan\_Trajectory}($X_{t},\mathcal{S},\mathcal{A},T,\mathcal{O},R,\gamma$)

\textbf{if} $q=$ \emph{null}\algoIndent{$q\leftarrow$ \textsc{Query\_Selector}($X_{t},\mathcal{O},R$)

\textsc{Request\_Label}($I_{q}$)}

\textbf{endif}

$t\leftarrow t+1$}
\end{myAlgorithm}

\begin{myAlgorithm}{\textsc{Plan\_Trajectory}}
\label{alg:plan-trajectory}
\textbf{Input:} $X_{t},\mathcal{S},\mathcal{A},T,\mathcal{O},R,\gamma$

\textbf{Given:} $n$ \tcp*{Number of trajectories to test}

$\mathcal{T}\leftarrow$ \textsc{Generate\_Trajectories}($X_t, \mathcal{S},\mathcal{A}, T, n$)

\textbf{for} $i=1,\dots,n$:
\algoIndent{$\boldsymbol{s}_{i}$ $\leftarrow$ \textsc{Score\_Trajectory}($\mathcal{T}(i),O,R, \gamma$)}
$\boldsymbol{\tau}\leftarrow\ \mathcal{T}(\argmax_{i} \boldsymbol{s}_{i})$

\textbf{return} $\boldsymbol{\tau}$
\end{myAlgorithm}

\subsection{Spatial Observation Model for Images}

A spatial observation model is required for adaptive path planning because the robot's reward is determined by what it observes, so to evaluate a candidate trajectory the robot must predict what it will observe along that trajectory. This should be possible because the semantic contents of natural images, such as terrain types and species present, often have strong spatial correlation~\cite{Flaspohler2017,Reiss2011}. However, these correlations are hard to model in the pixel space, where even nearly identical images can be made distant by effects like sensor noise and slight changes in illumination~\cite{Zhang2018}. Further, due to the high dimensionality of the image space, there are no spatial models with which it is computationally tractable to predict the image that would be observed in an unvisited location.

To overcome these challenges, the robot computes semantic representations of images in the space $\mathcal{Z}$, which is low-dimensional compared to the space of natural images $\mathbb{O}$. The robot builds a spatial observation model over semantic representations, denoted as
$\boldsymbol{Z}_{t}\left(\boldsymbol{x}\right): \mathbb{R}^3 \mapsto \mathcal{Z}$, trained using the observations $X_t = \{(x_i, I_i)\}_{i=1}^t$ and the semantic feature extractor $\boldsymbol{z}\left(I\right): \mathbb{O} \mapsto \mathcal{Z}$.
This approach requires that the semantic representations of two images $\boldsymbol{z}\left(I_1\right), \boldsymbol{z}\left(I_2\right)$ are similar (typically measured by Euclidean distance) if and only if the human-perceived similarity of $I_1$ and $I_2$ is high. Semantic representations derived from computer vision models developed for unsupervised natural image clustering, such as deep feature extractors~\cite{Romero2015,Zhang2018} and spatial topic models (STMs)~\cite{Wang2007,Girdhar2014a} have this property.

STMs such as BNP-ROST~\cite{Girdhar2019_ICRA, Girdhar2016} are a strong class of candidates for the spatial observation model because the priors they use to represent the spatial distributions of topics are smooth (spatially correlated) and the topic distributions they use to represent images have low dimensionality. The low-dimensionality of these representations is a critical requirement for learning the reward function from few examples; this is much more challenging with higher dimensional representations such as deep features~\cite{Zhang2018}. 




\subsection{Learning a Reward Model Online over Low Bandwidth}

We define the robot's reward to be the total number
of unique and interesting observations it has collected
\begin{align}
R\left(X_{t}\right) & =\sum_{i=1}^{t}\mathcal{I}\left(I_{i}\right) =\sum_{i=1}^{t}\left(Y_t\right)_i.
\end{align}
This can only be computed after the operator sees all images (i.e., after the mission). Since the robot models observations in the semantic space $\mathcal{Z}$,
trajectory planning requires it to estimate the reward as a function of the semantic field $\boldsymbol{Z}_t\left(\boldsymbol{x}\right)$.
For this, the robot learns a model $g_{\theta}:\mathcal{Z}\mapsto\left[0,1\right]$
\begin{equation}
R\left(\boldsymbol{x}\right)\approx g\left(\boldsymbol{Z}_t\left(\boldsymbol{x}\right); \theta\right),
\end{equation}
where $\theta$ is a set of parameters for the model family.
Recall that $L_t$ is the set of labeled image indices at time $t$, and let $D_t = \{(I_i, \left(Y_t\right)_i)\}_{i\in L_t}$ be the corresponding training set. We choose $\theta$ to minimize the cross-entropy loss $\mathcal{L}$ on $D_t$,
resulting in the final reward model
\begin{align}
R\left(\boldsymbol{x};D_t\right) & \approx g\left(\boldsymbol{Z}\left(\boldsymbol{x}\right); \theta_{D_t}^\star\right)\\
\theta_{D_t}^{\star} & =\argmin_{\theta}\sum_{(I,y)\in D_t} \mathcal{L}\left(y, g\left(\boldsymbol{z}\left(I\right); \theta \right)\right).
\end{align}

The number of labeled examples that a model must be trained on in
order to generalize well is proportional to the sample complexity of the
model family \cite{Mitzenmacher2016}, and for simple models (e.g., logistic regression) the sample complexity is typically linear in the number of input dimensions~\cite{Ng2002}. Thus, it is desirable to jointly pick a semantic representation
and a model $g_{\theta}$ such that the total number of examples required
to train $g_{\theta}$ is less than the number of examples
that can be labelled during the mission. This further motivates the use of BNP-ROST~\cite{Girdhar2016} as the semantic feature extractor, since the dimensionality of its semantic representation grows as $\log t$, logarithmic in the number of images $t$, while the number of labelled images grows linearly at $\frac{t}{n}$, where $n\ge1$ is set by the bandwidth constraint.
Thus, when using BNP-ROST
in combination with a simple reward model, then the training process for $g_{\theta}$ is expected to quickly converge to good parameters $\theta$, even with few training examples.

\subsection{Query Selection for Low Bandwidth Reward Learning}


When the robot observes novel phenomena, it needs to query the operator's interest in collecting more observations of the phenomena. 
The only type of query the robot can perform in an unknown environment is sending an image to the operator and receiving an interest label in return; the operator cannot determine their interest in an image from the image's semantic representation, and does not have access to enough information to advise the robot on the optimal policy.
This is a unique challenge 
%
for active learning.

\section{Online Active Reward Learning for POMDPs}\label{sec:active-learning}

Here we will consider active learning strategies to learn the parameters of a POMDP reward model online. We denote the set of unlabelled image indices at time $t$ as $\mathcal{U}_{t}$, and the active learning metric as $h\left(\boldsymbol{z}\right)$, such that the next image to request a label for is chosen as
\begin{equation}
i^{\star}=\argmax_{i\in\mathcal{U}_{t}}h\left(\boldsymbol{z}\left(I_{i}\right)\right).
\end{equation}

\subsection{Non-Adaptive Query Selection}

The simplest approaches to selecting images to be labelled do not depend on $\boldsymbol{z}$, and thus are good baselines to consider. \textit{Random} selection chooses unlabelled observations uniformly at random. \textit{Uniform} selection instead chooses every $n^\text{th}$ image, where $n$ is determined by the bandwidth constraint.

\subsection{Informative Query Selection}

Informative query selection involves defining some uncertainty metric on the model, and choosing to label the observation which results in the greatest reduction of uncertainty. There are many query selection strategies that fall into this category and are effective at learning a function in few examples \cite{Yang2018}. A common uncertainty metric for classification problems is entropy, where the highest entropy values occur when an observation is on a decision boundary. A widely-used approach to informative query selection is ``uncertainty sampling'', which typically means picking the observation with the maximum entropy \cite{Yang2018}
\begin{align}\label{eq:entropy}
h_\text{Entropy}\left(\boldsymbol{z}; \theta_{D_t}^\star\right) &=\mathbb{H}\left[g\left(\boldsymbol{z}; \theta_{D_t}^\star\right)\right].
\end{align}
An issue with uncertainty sampling is that labeling the most uncertain observation might not have much effect on the model parameters $\theta$ -- if the model parameters do not change, then the model performance does not increase. This suggests maximizing ``error reduction'' \cite{Yang2018} instead
\begin{align}
    \label{eq:info-gain}h_\text{Info}\left(\boldsymbol{z}\right) &= h_\text{Entropy}\left(\boldsymbol{z}; \theta_{D_t}^\star\right) - \mathbb{E}_{D^\prime_t \mid D_t
    }\left[h_\text{Entropy}\left(\boldsymbol{z}; \theta_{D_t^\prime}^\star\right) \right]\\
    &D_t^\prime = D_t  \cup (\boldsymbol{z}, y)\nonumber\\
    &p(D^\prime_t \mid D_t) \approx  g(\boldsymbol{z}; \theta_{D_t}^\star)\nonumber.
\end{align}
This \textit{Information Gain} query selection method prioritizes labeling an observation by how much a new label $y$ is expected to reduce the entropy of similar future observations. This should maximize the rate at which entropy is reduced and thus the rate at which the reward function is learned.

\subsection{Regret Minimizing Query Selection}

Here we introduce a novel \textit{Regret} minimizing query selector that focuses on identifying labels that maximize the expected reward collected during the mission, rather than information gained about the reward function. Regret is typically defined for POMDPs as the difference in utility between the chosen action and the true optimal action based on complete information. To our knowledge, this is the first work that compares a regret-based heuristic with information-theoretic heuristics in online active learning.   

Suppose that the robot is considering a finite set of trajectories $\mathcal{T} = \{\tau_i\}_{i=1}^{N_\tau}$: it uses the observation model to predict what it will observe along each trajectory $\tau$, predicts each trajectory's reward, and finally chooses the one with the highest reward (see Algorithm \ref{alg:plan-trajectory}). However, given limited training data, the robot has significant uncertainty in the predicted rewards and thus is unlikely to have chosen the true optimal trajectory. This motivates a question for each unlabeled image: if this image were labeled, would the robot have chosen a different trajectory? If the answer is yes, then it must mean that, given this additional label, a different trajectory would be predicted to have greater reward and thus the robot would ``regret'' not knowing the label. If it is no, then the robot would have no immediate regret for not knowing it. We formalize this in the following objective:
\begin{align}
    \label{eq:regret}h_\text{Regret}\left(\boldsymbol{z}\right) &= \mathbb{E}_{D_t^\prime \mid D_t}\left[R(\tau^\star_{D_t^\prime}; D_t^\prime) - R(\tau^\star_{D_t}; D_t^\prime)\right]\\
    R(\tau; D_t) &= \sum_{\boldsymbol{x}\in\tau} g(\boldsymbol{Z}(\boldsymbol{x}); \theta^\star_{D_t})\\
    \tau_D^\star &= \argmax_{\tau\in \mathcal{T}} R(\tau; D) 
\end{align}
Equation \ref{eq:regret} may be interpreted as the expected reward increase (regret decrease) given a label for $\boldsymbol{z}$. An approach to computing $h_\text{Regret}$ is presented in Algorithms \ref{alg:regret-query} and \ref{alg:regret}.

\begin{myAlgorithm}{Regret-Based Query Selection}
\label{alg:regret-query}
\textbf{Given:} $X_{t},\mathcal{S},\mathcal{A},T,\mathcal{O},R,\gamma$

\textbf{Input:} $\mathcal{U}_t$ \tcp*{Set of unlabeled image indices}


$\tau_0$ $\leftarrow$ \textsc{Plan\_Trajectory}($X_{t},\mathcal{S},\mathcal{A},T,\mathcal{O},R,\gamma$)

\textbf{foreach} $i \in \mathcal{U}_t$:\algoIndent{$\boldsymbol{z}\ \leftarrow$ \textsc{semantic\_representation}($I_{i}$)

$y_{\text{pred}}\leftarrow$ \textsc{Predict\_Reward}($\boldsymbol{z}$)

$r_{0}$ $\leftarrow$ \textsc{Compute\_Regret}($\tau_0, \boldsymbol{z}$, 0)

$r_{1}$ $\leftarrow$ \textsc{Compute\_Regret}($\tau_0, \boldsymbol{z}$, 1)

regret$_{i}$ $\leftarrow$ $y_{\text{pred}}r_{1}+\left(1-y_{\text{pred}}\right)r_{0}$} 

\textbf{return} $\argmax_{i\in \mathcal{U}_t}$ regret$_{i}$
\end{myAlgorithm}
\begin{myAlgorithm}{\textsc{Compute\_Regret}}
\label{alg:regret}
\textbf{Given:} $X_{t},\mathcal{S},\mathcal{A},T,\mathcal{O},R,\gamma$

\textbf{Input}: $\tau_0, \boldsymbol{z}, y$ \tcp*{Reference trajectory, observation to label, and temporary label}

\textsc{Add\_Temporary\_Label}($\mathcal{O}, \boldsymbol{z}, y$)

$\tau^\star$ $\leftarrow$ \textsc{Plan\_Trajectory}($X_{t},\mathcal{S},\mathcal{A},T,\mathcal{O},R,\gamma$)

$s^\star \leftarrow$ \textsc{Score\_Trajectory}($\tau^\star,\mathcal{O},R,\gamma$)

$s_0 \leftarrow$ \textsc{Score\_Trajectory}($\tau_0,\mathcal{O},R,\gamma$)

\textsc{Remove\_Temporary\_Label}($\mathcal{O}, \boldsymbol{z}$)

\textbf{return} $(s^\star - s_0)$ \tcp*{Regret given the temp label}
\end{myAlgorithm}

\begin{figure*}
\begin{subfigure}{.4\linewidth}
    \centering
    \includegraphics[width=.665\linewidth]{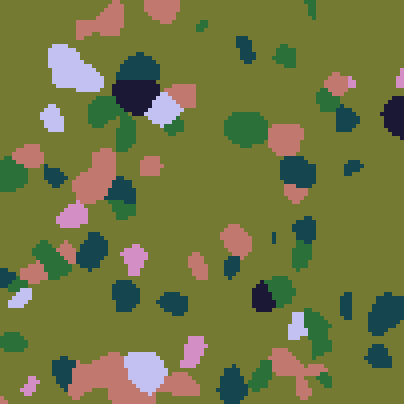}\\
    \vspace{0.25em}
    \includegraphics[width=.665\linewidth]{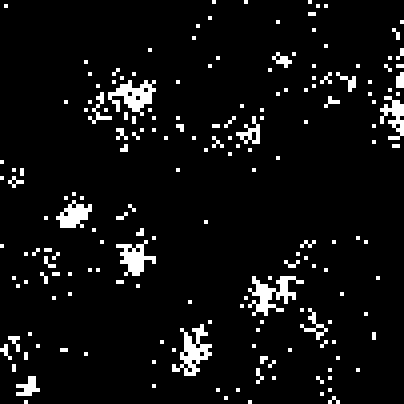}
    \subcaption{\label{subfig:map}Top: A topic map where each location is described by the semantic representation $\boldsymbol{z}^{(i,j)} \in \boldsymbol{\Delta}^8$. The color of each pixel indicates the largest component of $\boldsymbol{z}^{(i,j)}$. Bottom: The reward at each location is randomly sampled as $R^{(i,j)} \sim \text{Bernoulli}(\boldsymbol{p}^T\boldsymbol{z}^{(i,j)})$, where $\boldsymbol{p} \in [0,1]^k$ represents how ``interesting'' each component of $\boldsymbol{z}$ is. Here, the pink and black topics are most interesting.}
\end{subfigure}
\hfill
\begin{subfigure}{.57\linewidth}
    \centering
    \includegraphics[width=.46\linewidth]{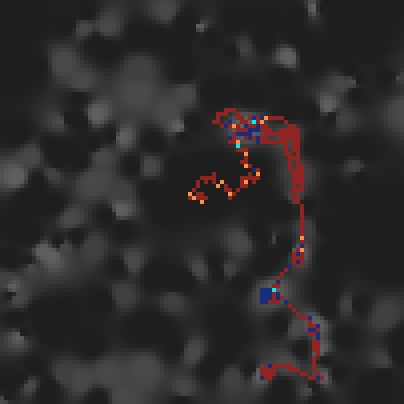}
    \includegraphics[width=.46\linewidth]{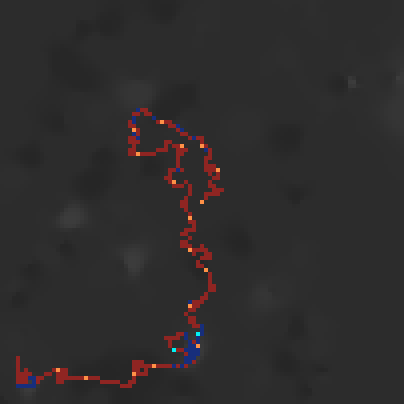}\\\vspace{0.25em}
    \includegraphics[width=.46\linewidth]{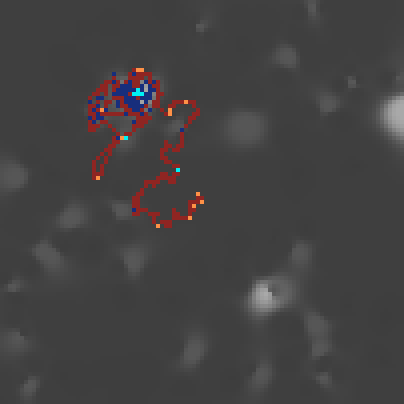}
    \includegraphics[width=.46\linewidth]{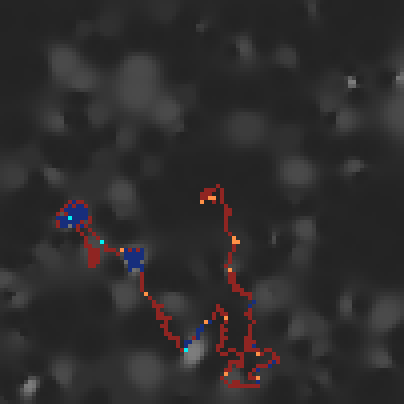}
    \subcaption{\label{subfig:trajectories}Best viewed on a screen. Sample trajectories followed by robots starting at the center of the map in (\subref{subfig:map}) with different query selectors. Along each trajectory, red-orange pixels correspond to no reward, and blue pixels to reward. Bright orange/blue pixels represent observations for which the query selector requested the label. The greyscale background intensities represent $g(Z(x);\theta^\star_D)$: reward estimates of observations at each location, based on all labeled samples. Query Selectors: (top row) Random, Uniform; (bottom row) Info Gain, Regret.}
\end{subfigure}\\
\begin{subfigure}{\linewidth}
    \centering
    \includegraphics[width=.675\textwidth]{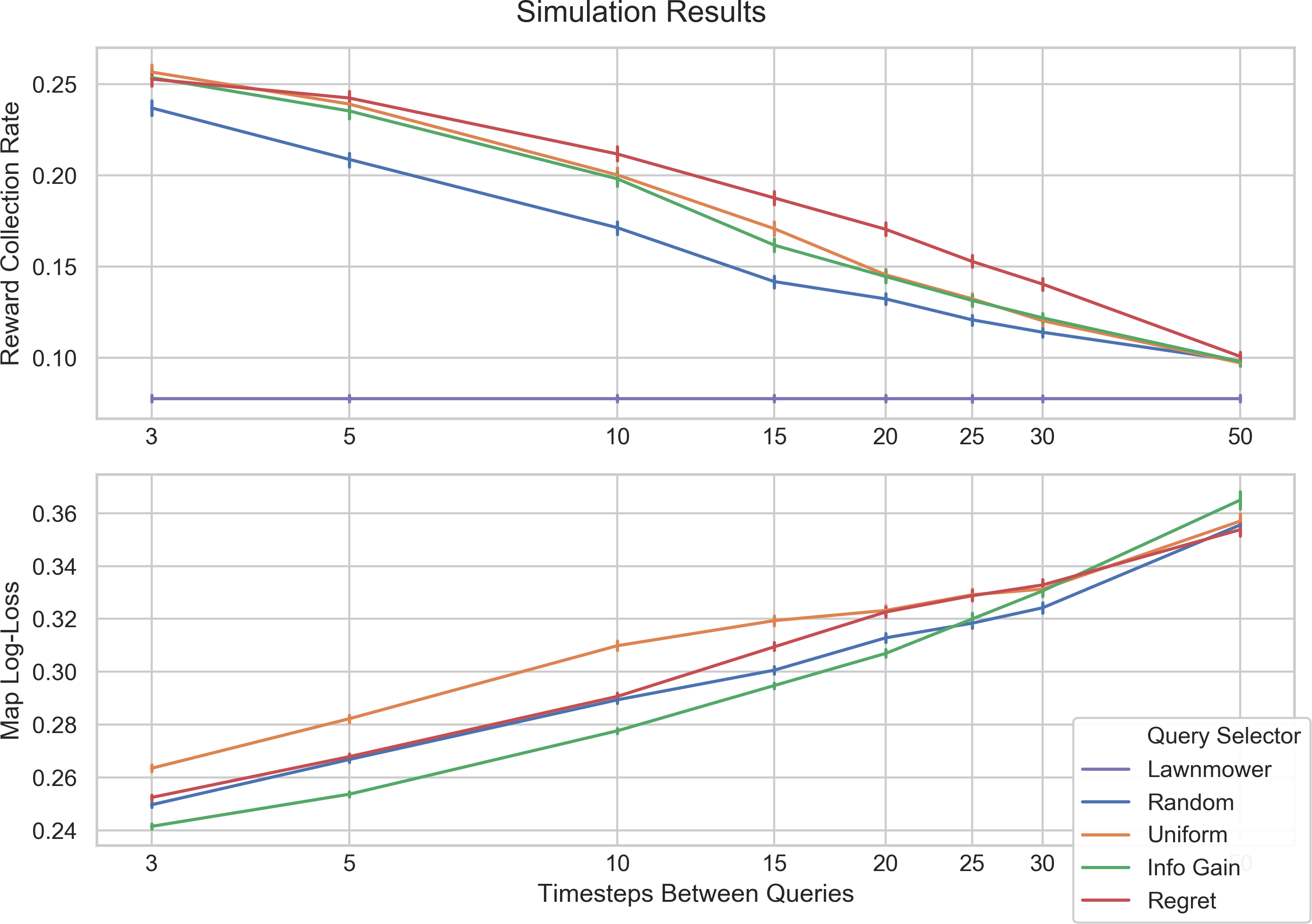}
    \subcaption{\label{subfig:exp-a-results}A comparison of the query selector performance for different bandwidth availability; the x-axis represents labeling period (time between making a call to \textsc{Request\_Label} and \textsc{Label\_Ready} returning true in Algorithm \ref{alg:corobotic-exploration}), which is inversely proportional to bandwidth. Each datapoint represents the mean of 1080 simulations (36 trials on 30 unique maps) and bars represent the 68\% confidence bound of the mean. Top: The mean amount of reward collected by each robot per unit time (higher is better). Lawnmower is not a query selector, but rather represents the mean reward collected by 8 preplanned boustrophedonic trajectories \cite{Choset1998} that each start at the center of the map and move towards a corner. Bottom: The mean cross-entropy loss between the ground truth interest maps, as in (\subref{subfig:map}), and the corresponding robots' predictions of the reward at each location, as in (\subref{subfig:trajectories}), at the end of each simulation (lower is better).}
\end{subfigure}
    \caption{Stages of the simulation procedure, and performance comparison of the query selectors on fully simulated data.}
    \label{fig:simulated_demo}
\end{figure*}

\section{Experiments}
\label{sec:experiments}

\begin{figure}
    \centering
    \includegraphics[width=.32\linewidth]{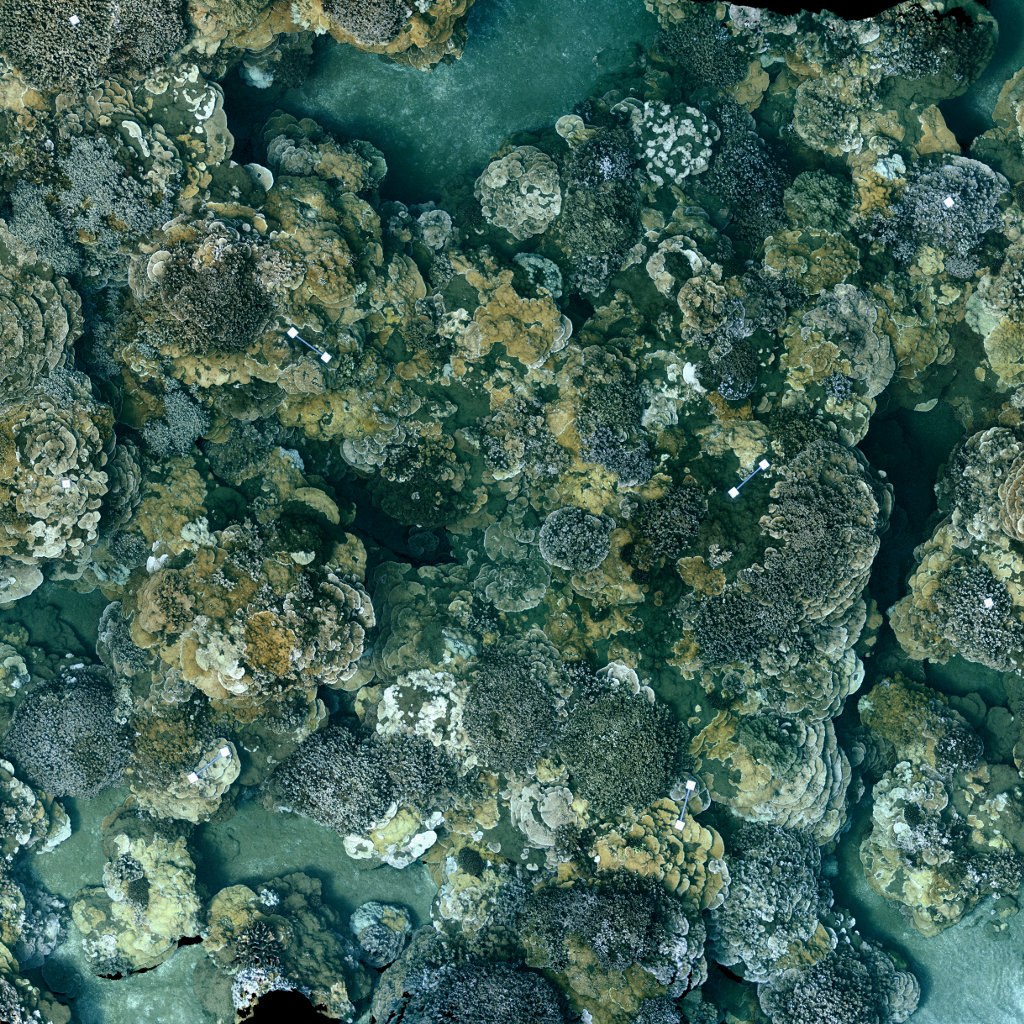}
    \includegraphics[width=.32\linewidth]{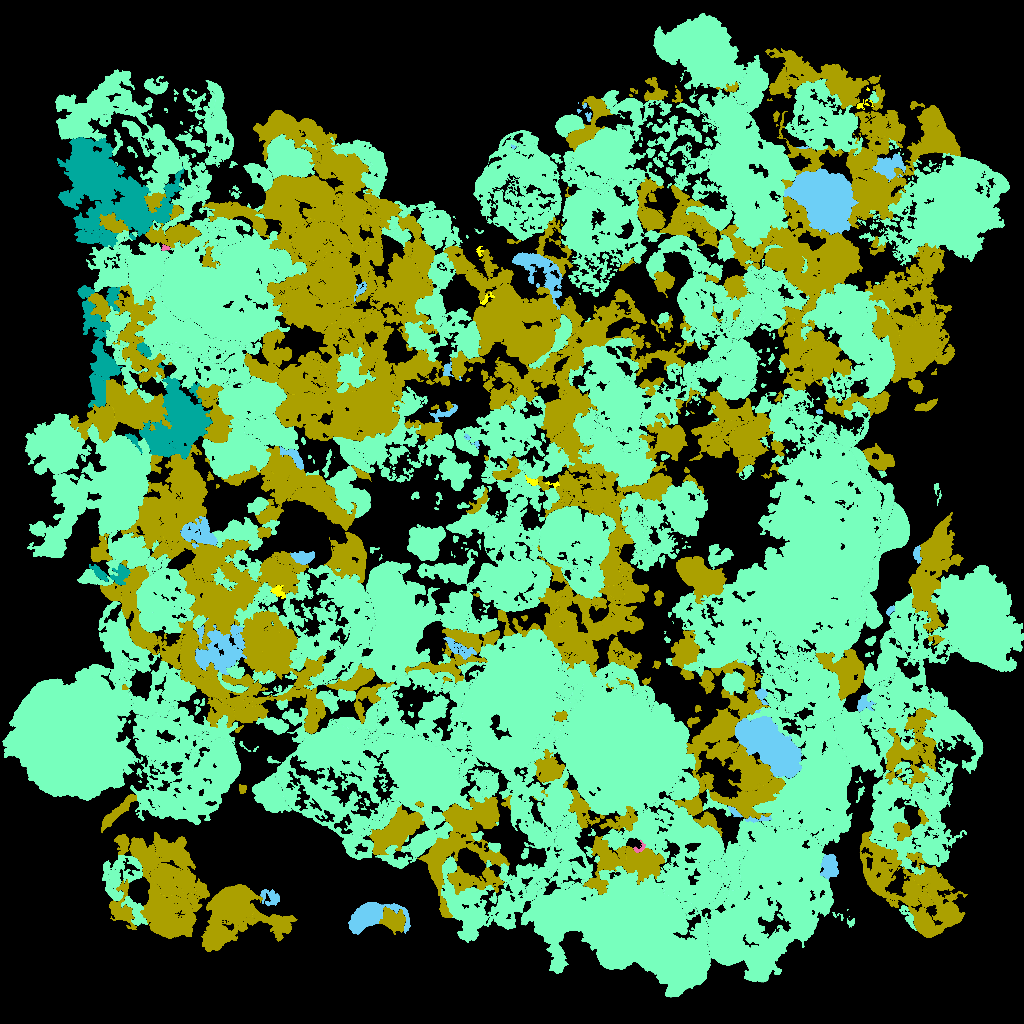}
    \includegraphics[width=.32\linewidth]{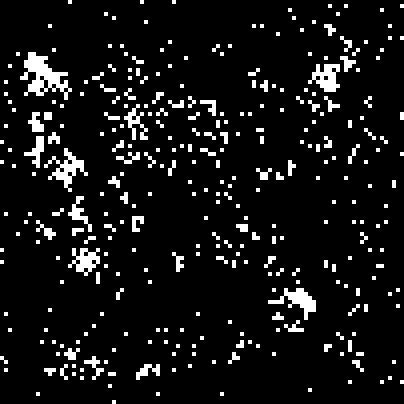}
    \caption{Left: A crop of the KAH\_2016\_3 photomosaic image from the 100 Islands Challenge \cite{Smith2016}, showing a coral reef near Kaho'olawe. Center: The photomosaic annotations where each color represents an expert label \cite{Smith2016}. 
    Right: One of 30 unique interest maps generated (cf. \autoref{subfig:map}).}
    \label{fig:kah_demo}
\end{figure}

We evaluate the proposed query selection techniques through two experiments, each simulating the co-robotic exploration task with various bandwidth constraints. The first experiment (see Figure \ref{fig:simulated_demo}) used 30 artificial ``topic maps'' (cf. \cite{Girdhar2019_ICRA}) created by randomly generating Voronoi partitions of a 100\texttimes100 image, assigning each cell a topic label, and then assigning each pixel's topic distribution as a distance-weighted mean over cell labels. This produced continuous topic maps with topics in varying concentrations, and each one was associated with a unique interest map (see Figure \ref{subfig:map}). In the second experiment, a single topic map was derived from the expert annotations of an actual coral reef image, and 30 interest maps were generated for it (see Figure \ref{fig:kah_demo}).
The procedure for both experiments was:
\begin{enumerate}
    \item Generate a map of topic distributions $\boldsymbol{z}(\boldsymbol{x}) \in \boldsymbol{\Delta}^d$ which represent the observations at each location $\boldsymbol{x}$;
    \item Generate an interest profile $\boldsymbol{p} \in [0,1]^d$ so that $p = \boldsymbol{p}^T\boldsymbol{z}$ is the probability that the operator is interested in an observation with feature representation $\boldsymbol{z}$;
    \item Generate a binary ``interest map'' by sampling $R(\boldsymbol{x}) \sim \text{Bernoulli}(p(\boldsymbol{z}(\boldsymbol{x}))$ at each location $\boldsymbol{x}$ in the topic map;
    \item For each bandwidth limitation and each query selection algorithm: perform 36 rollouts of \autoref{alg:corobotic-exploration} for a simulated robot making reward queries according to the bandwidth limitation and query selector
\end{enumerate}

Each rollout in step (4) had a duration of 300 timesteps; robot movement was one pixel per timestep and bandwidth constraints were simulated by changing the number of timesteps for a label to be received after being requested. State transitions and observations were deterministic and noiseless. The robot started with no training data and used logistic regression (from \cite{scikit-learn}) as its reward model. Trajectories were generated by randomly sampling sequences of 5 motion primitives.\footnote{The primitives were 13 straight lines, each 5 units long and at angles spaced uniformly between -135$^\circ$ to 135$^\circ$ from the robot's current direction.} 50 trajectories were generated at each timestep and scored using the sum of the predicted rewards along the trajectory, less the scores of locations already visited. The highest scoring trajectory was followed.

\section{Results \& Discussion}

\begin{figure}
    \centering
    \includegraphics[width=\linewidth]{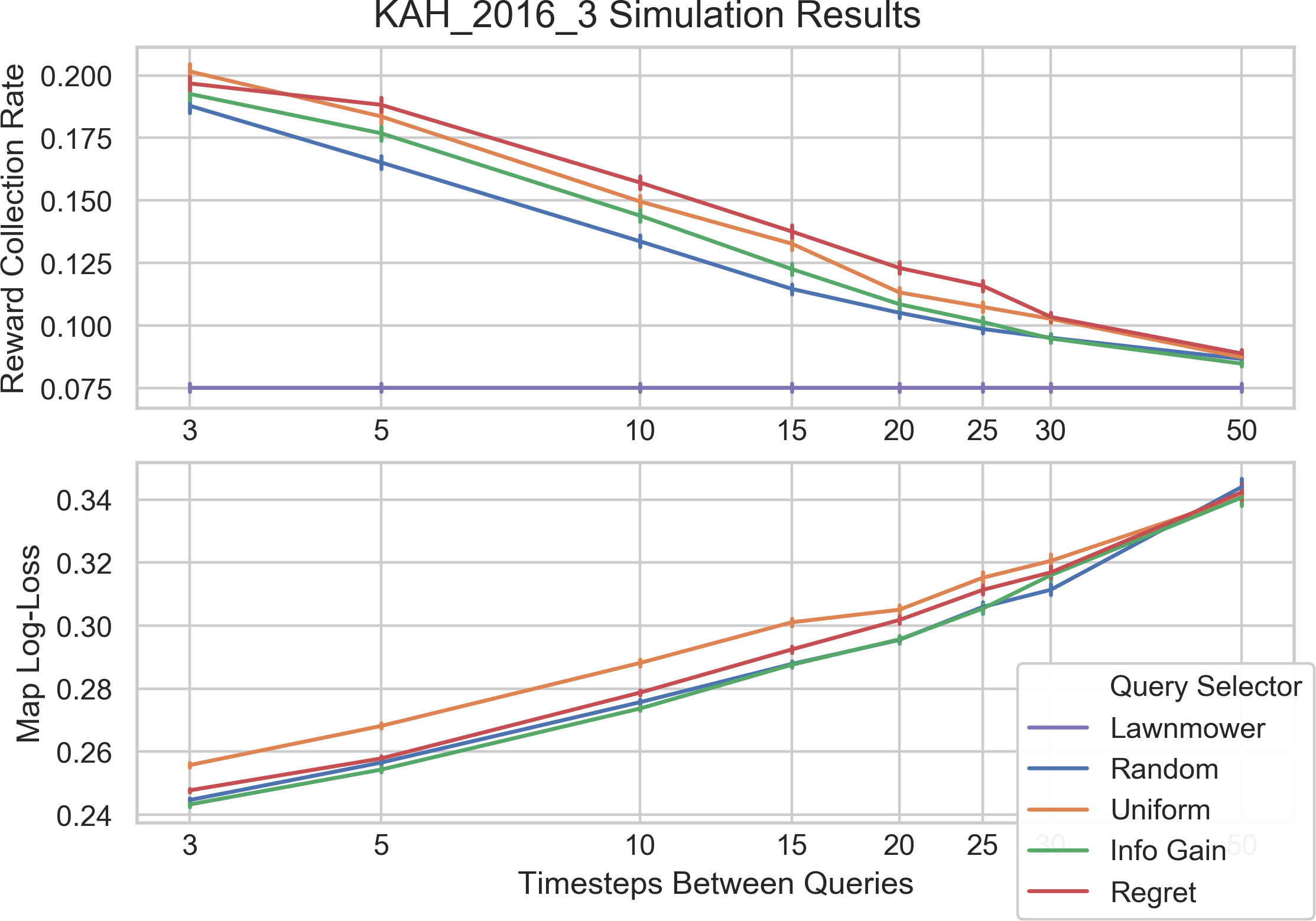}
    \caption{The Regret query selector continues to outperform the other active learning heuristics when the topic map is derived from a real image (see \autoref{fig:kah_demo}).}
    \label{fig:kah_results}
\end{figure}

We compared the Random, Uniform, Information Gain, and Regret query selectors described in Section \ref{sec:active-learning} over a total of 69120 simulations; the mean reward collection rates and interest map prediction losses for each experiment are presented in Figures \ref{subfig:exp-a-results} and \ref{fig:kah_results}. The Regret query selector matches, or outperforms, every other selection criterion at collecting reward, at any bandwidth availability, in these simulation configurations. The relative gains of non-random query selection are smaller when the time between queries is short (high-bandwidth) and thus almost every image is labeled, or when it is so long (low-bandwidth) that the robot barely learns anything before the mission ends. The results also demonstrate the vast improvement of autonomous exploration over preplanned trajectories: the adaptive planners collected up to 29.7\% more reward at very low bandwidth, and up to 230\% more reward at high bandwidth.

The regret-based method did not learn the reward function as well as the information gain query selector, based on its higher map log-loss. This exemplifies the difference in the design criteria: the information theoretic criterion focuses on useful labels for learning a function, which is appropriate for active reward learning \textit{offline}, during training. The regret criterion instead optimizes for the robot's reward, making it better suited for \textit{online} active reward learning, which describes our usage of queries during a live mission. 


\section{Conclusions and Future Work}

The Co-Robotic Visual Exploration POMDP provides a structured approach to managing human-robot collaboration and high-dimensional observation spaces in autonomous science. We provide general principles for choosing the POMDP's observation model, reward model, and active learning criterion, and demonstrate that the novel Regret-based active learning criterion can greatly improve the amount of reward collected. Some next steps are: exploring spatial observation models capable of longer-range topic prediction (e.g. \cite{SanSoucie2020Gaussian-DirichletObservations}), extending the reward model and active learning formulation to non-binary rewards, and using higher-fidelity simulations and field deployments to better understand the performance increases that can be achieved in real-world autonomous exploration.

\bibliographystyle{IEEEtran}

\ifthenelse{\boolean{arxiv}}{

}{
\bibliography{IEEEabrv,stewart}
}

\end{document}